\def\year{2025}\relax
\documentclass[letterpaper]{article}
\usepackage{aaai}
\usepackage[round]{natbib}
\usepackage{times}
\usepackage{helvet}
\usepackage{courier}
\usepackage{enumerate}
\usepackage{multirow}
\usepackage{indentfirst} 
\usepackage[]{graphicx}
\usepackage{epsfig}
\usepackage{epsf}
\usepackage{placeins}
\usepackage{float}
\begin{document}
\title{MMCE: A Framework for Deep Monotonic Modeling of Multiple Causal Effects}
\author{
  Juhua Chen\textsuperscript{1}, Karson shi\textsuperscript{2}, Jialing He\textsuperscript{3}, North Chen\textsuperscript{4}, Kele Jiang\textsuperscript{5}
}
\maketitle
\footnotetext[1]{Email: chenjuhua.cjh@alibaba-inc.com}
\footnotetext[2]{Email: karson@alibaba-inc.com}
\footnotetext[3]{Email: hjl391803@alibaba-inc.com}
\footnotetext[4]{Email: north.chen@alibaba-inc.com}
\footnotetext[5]{Email: kele.jiang@alibaba-inc.com}
\begin{abstract} 
\begin{quote}
When we plan to use money as an incentive to change the behavior of a person (such as making riders to deliver more orders or making consumers to buy more items), the common approach of this problem is to adopt a two-stage framework in order to maximize ROI under cost constraints. In the first stage, the individual price response curve is obtained. In the second stage, business goals and resource constraints are formally expressed and modeled as an optimization problem. The first stage is very critical. It can answer a very important question. This question is how much incremental results can incentives bring, which is the basis of the second stage. Usually, the causal modeling is used to obtain the curve. In the case of only observational data, causal modeling and evaluation are very challenging. In some business scenarios, multiple causal effects need to be obtained at the same time. This paper proposes a new observational data modeling and evaluation framework, which can simultaneously model multiple causal effects and greatly improve the modeling accuracy under some abnormal distributions. In the absence of RCT data, evaluation seems impossible. This paper summarizes three priors to illustrate the necessity and feasibility of qualitative evaluation of cognitive testing. At the same time, this paper innovatively proposes the conditions under which observational data can be considered as an evaluation dataset. Our approach is very groundbreaking. It is the first to propose a modeling framework that simultaneously obtains multiple causal effects. The offline analysis and online experimental results show the effectiveness of the results and significantly improve the effectiveness of the allocation strategies generated in real-world marketing activities.
\end{quote}
\end{abstract}
\section{1 Introduction}
Online food delivery is experiencing explosive growth and the service is becoming increasingly popular. Food delivery platforms aim to provide high-quality and stable services to customers and restaurants. To solve this problem, operation  managers provide a certain amount of special funds to encourage crowd-sourced riders to complete more orders. This problem can also be solved using a two-stage framework~\citep{Gupta_Steenburgh_2008}. In the first stage, individual price response curves are obtained. In the second stage, business objectives and resource constraints are formally expressed and modeled as an optimization problem.

Changes in user responses caused by different incentives can be regarded as estimation of the individual treatment effect(ITE)~\citep{zhang2021unified}. It is often called uplift. One solution to obtain unbiased estimation is to use a completely random assignment strategy ~\citep{Sill_Abu-Mostafa_1996} to collect a large amount of unbiased data. Due to limited budgets, this method is not practical and is sometimes not allowed(it is harmful to the rider experience). In the case of only observational data, causal modeling and evaluation are very challenging.

There are a lot of confounding factors in the observational data. Figure 1 shows the relationship between the incentive amount and the number of completed orders under random data and biased data. Under random data, the number of completed orders increases with the increase of incentives. This  is consistent with our cognition. In the observational data, the number of completed orders decreases with the increase of incentives sometimes. This happens because the previous allocation strategy was intervened. High-active riders are allocated a small amount of incentives, while low-active riders are the opposite. The response model trained with observational data will overestimate the number of completed orders for small incentives and underestimate the number of completed orders for large incentives. The most serious is that even the basic monotonic trend that the number of completed orders increases with the increase of incentives cannot be met.
\begin{figure*}[h]
\centering
\noindent \includegraphics[scale=0.54]{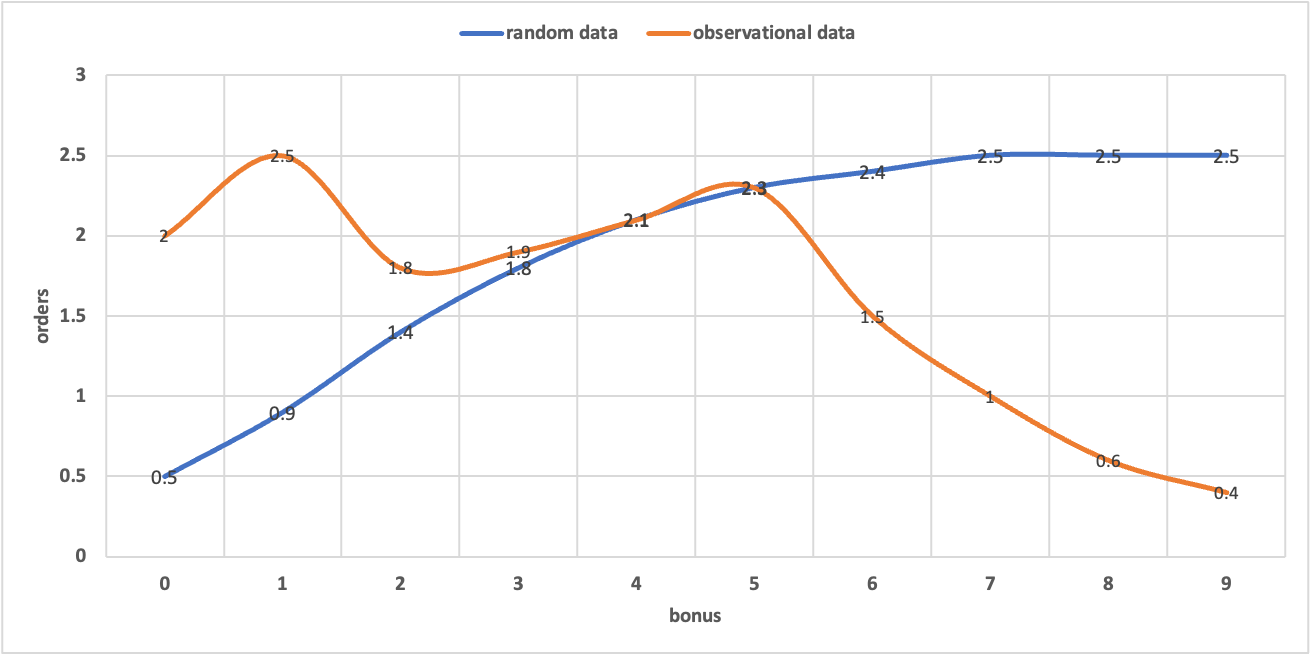}
\caption{The relationship between the incentive amount and the number of completed orders under random data and observational data}
\label{fig:method}
\end{figure*}

There are many methods for modeling with observational data. One method requires fusion modeling with a small amount of RCT data. Another method does not require RCT data and uses observational data for modeling, but the evaluation depends on RCT data. This article is based on a scenario where there is no RCT data at all and only uses observational data for modeling. The evaluation is conducted using cognitive testing, and quantitative evaluation indicators for priori are established.

In order to address the existence of confounding factors in observational data, many modeling methods have been proposed. One type of method puts the idea of propensity score into the model structure. This idea has a strong dependence on the estimated accuracy of the propensity score. The propensity score cares not about the result itself but its ability to balance covariates. In the case of massive data in the industry, due to the existence of various personalized scenarios, the allocation strategy also has certain differences, and the propensity score method may fail. One type of method is to modify the model structure directly to meet this monotonic prior requirement because it is found that the observational data does not even meet the basic monotonic prior. This method is simple to implement, highly interpretable, and very extensible. This method is adopted in this paper.

In the method of modeling using observational data, generally only one price response curve is output. However, in some business scenarios, the response results have obvious sequence characteristics, and the distribution of the response results is also relatively distorted. Directly modeling a causal effect has a very large deviation. This paper proposes a framework for multi-stage causal modeling based on large-scale observational data without RCT data at all, which generates multiple causal effects at the same time. In this framework, the business prior monotonicity will be satisfied, and we call it the framework for Deep Monotonic Modeling of Multiple Causal Effects(MMCE).

In order to verify the effectiveness and efficiency of the framework, we conducted an offline evaluation on the real data set  and conducted an online A/B test on the crowdsourcing platform . The online experiment shows that the actual ROI of our proposed framework has increased by 12\%. Under the guidance of the MMCE framework concept, we continue to implement a better version, which is expected to have a greater effect.

To summarize, our main contributions are:
\begin{itemize}
\item We are the first to propose a framework for multi-stage causal modeling based on large-scale observational data to generate multiple causal effects at the same time.
\item The network structure is highly scalable.
\item This paper summarizes three basic priors to illustrate the necessity and feasibility of qualitative evaluation of cognitive testing. At the same time, this paper innovatively proposes under what conditions observational data can be considered as an evaluation data set.
\end{itemize}
\section{2 Related Works}
\textbf{Causal Modeling of Observational Data} There are many methods in this field. When the data collection process is not random, we use the inverse propensity score (PS) ~\citep{austin2011introduction} to weight the sample data. PS is also used in meta-learning methods such as X-learner ~\citep{kunzel2019metalearners}/R-learner ~\citep{nie2021quasi}/DR-learner ~\citep{kennedy2020optimal}.
Due to the flexibility of neural networks, deep learning methods for estimating causal effects have also surged recently. DragonNet ~\citep{shi2019adapting}/VCNet ~\citep{nie2021vcnet} and other methods also use the idea of PS. Representative methods for continuous treatment include DRNet ~\citep{schwab2020learning} and VCNet ~\citep{nie2021vcnet}. It is generally believed that DRNet is a special case of VCNet, which directly obtains the dose-response curve. In terms of network structure, Euen ~\citep{ke2021addressing} corresponds to two sub-networks for natural quantity and incremental quantity respectively. The sum of the results of the two sub-networks equals the final observation quantity, which is similar to the idea of the internal network structure of MMCE.
DESCN ~\citep{zhong2022descn} is an end-to-end multi-task cross-network model that hopes to alleviate the problems of treatment bias and sample imbalance in observation data modeling, which is also similar to the idea of the internal network structure of this paper.

\textbf{Monotonicity Modeling} Monotonic architecture is achieved by constructing a neural architecture that guarantees monotonicity. For example, the representative method is DLN~\citep{you2017deep}. It is highly complex and has some limitations that sometimes make it impossible to obtain the required training results~\citep{runje2023constrained}.
Monotonicity can be enforced during training by modifying the loss function or adding regularization terms ~
\citep{Sill_Abu-Mostafa_1996}~\citep{Gupta_Shukla_Marla_Kolbeinsson_Yellepeddi_2019}. These methods are easy to implement and can be used with any neural network architecture, but they cannot strictly guarantee the monotonicity of the training model.
Another method is to directly add a monotonic layer to the network structure. The monotonic layer is an implementation of the monotonic function, such as~\citep{shen2021framework}. This method is easy to implement and can be used with any neural network architecture. It can be flexibly modified according to business priors and is highly interpretable. This method is used in MMCE. This paper provides multiple candidate sets of monotonic functions for the monotonic layer, see the appendix.
\section{3 Methodology}
\subsection{3.1 Deep monotonic model paradigm for causal effect estimation}
Different monotonic functions can be used, and the specific deep neural network can be different. The preprocessing of input data can also be different, but the general deep monotonic model paradigm for causal effect estimation is shown in Figure 2.
\begin{figure*}[h]
\centering
\noindent \includegraphics[scale=0.54]{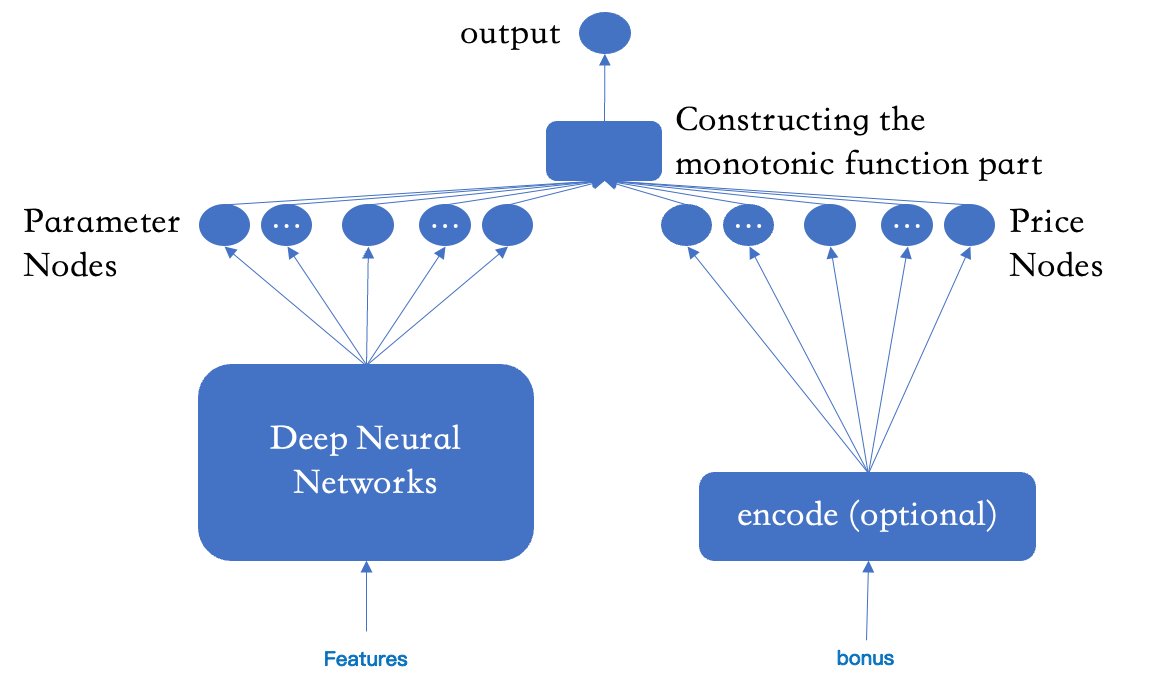}
\caption{General deep monotonic  model paradigm for causal effect estimation}
\label{fig:method}
\end{figure*}
\begin{itemize}
\item Deep neural network can select a suitable deep network architecture. It can be reasonably selected according to the business characteristics. In this part, similar to the idea of using PS in DragonNet~\citep{shi2019adapting}/VCNet~\citep{nie2021vcnet}, it is used to learn the distribution of intervention level t under given feature X. Using $E(Y|\pi(t|x),T=t)$ to estimate $ E[Y|X,T=t]$ helps to remove noise and extract useful information from features. This part can also use this structure.
\item The number of parameter nodes depends on the number of parameters of the monotonic function you choose.
\item Encoding part depends on whether the monotonic function you choose needs encoding and the encoding method to determine the number of price nodes.
\item Constructing monotonic function part is to implement the monotonic function you choose.
\end{itemize}
The entire data(various data preprocessing is not discussed in this article) is used to train the network, and information is fully shared. More importantly, the parameters to be learned are the output part of the deep neural network. During online prediction, each data can get its own parameter, which can be considered to be a concept similar to the variable coefficient in VCNet~\citep{nie2021vcnet}.

\subsection{3.2 The simplest s-shaped network}——
If the monotonic function uses a simple s-shaped function, like 
$y = \frac{D}{1+e^{-a*x + b }}$.
The deep neural network uses MLP, and the number of parameter nodes is 2, namely node a and node b, corresponding to the parameters a and b in the monotonic function. No encoding is required at this time, and the PS part is also omitted. The number of price nodes is 1, corresponding to the original price information x. The monotonic function part is constructed according to the calculation formula of the monotonic function. The network structure is shown in Figure 3.
\begin{figure*}[h]
\centering
\noindent \includegraphics[scale=0.44]{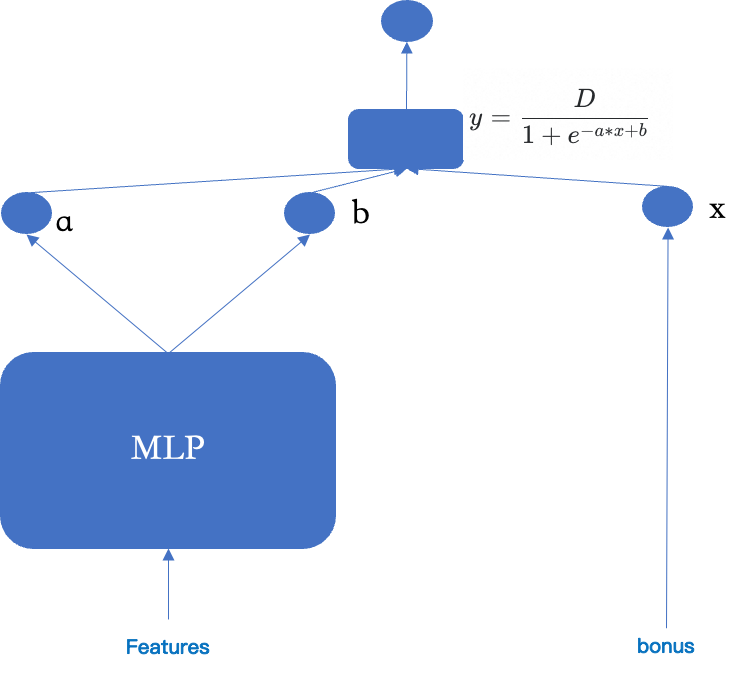}
\caption{The simplest s-shaped network structure}
\label{fig:method}
\end{figure*}

\subsection{3.3 Modeling Challenges}
Based on the previous deep monotonic model paradigm, we can continue to optimize the network structure to solve practical modeling challenges.
\begin{itemize}
\item Like "display then click" in the advertising system, the relationship between the result and the behavior sequence can be expressed as a multiplication structure.The formal formula is as follows.

$pCTCVR = pCTR * pCVR$

$pCTCVR = p(y=1,z=1|x)$

$pCTR = p(y=1|x)$

$pCVR = p(z=1|y=1,x)$

The ESMM~\citep{ma2018entire} structure is used to model the relationship between variables. Based on the user behavior sequence and multi-task learning ideas, it effectively solves the sample selection bias(SSB) and data sparsity(DS) problem. This problem is universal. For example, in the field of instant delivery , the number of orders completed by the rider can also be understood as whether the rider will  work (attendance rate) and how many orders he will complete if he begins to  work(completed orders after attendance). The formal formula is as follows.

$Orders = pAttendance * pPostAttendanceOrders$

$Orders = f(y|x) $

$pAttendance = p(t=1|x) $

$pPostAttendanceOrders = f(y|t=1,x) $

\item When the response in the price response curve is a continuous value and the long-tail distribution is very serious, the effect of direct modeling is very poor. Google also proposed ZILN\citep{wang2019deep} for this purpose, because the different loss functions of conventional regression modeling have assumptions about the distribution of input data, such as the mse loss  expects to satisfy the normal distribution. Splitting a deformed distribution into two relatively normal distributions can obtain more accurate results. On the other hand, in some cases, it is hoped that two causal effects can be obtained at the same time.
\item Based on the potential outcome framework, there is the following relationship: $\mu_t = \mu_c + \tau$, where $\mu_c$ represents the outcome of the control group, which can be understood as the result without intervention, and $\tau$ represents uplift, which is what we often call ITE or CATE, which can be understood as the incremental transformation brought about by intervention. In our business, we also call $\mu_c$ the natural quantity and $\tau$ the incremental quantity. For processes with only observational data and using monotonic networks to model response curves, there is also a choice of "separate modeling to obtain response curves" or "directly building response curves" for natural quantity and incremental quantity. Euen~\citep{ke2021addressing} explained the effectiveness of separate modeling.
\end{itemize}

Based on the previous analysis, we have multiple options for the modeling framework, as shown in Figure 4 below.
\begin{figure*}[h]
\centering
\noindent \includegraphics[scale=0.54]{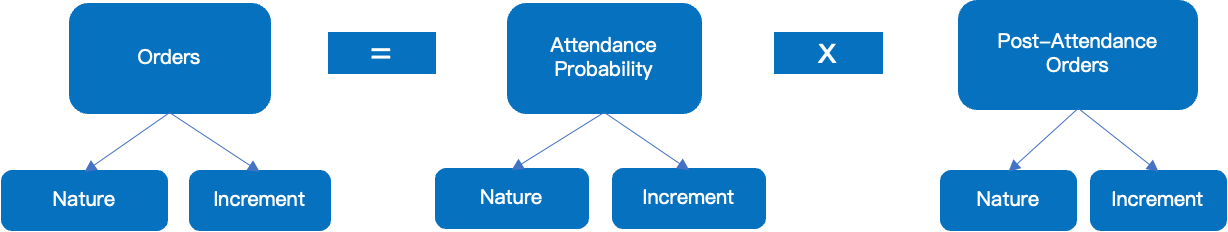}
\caption{Differenct types of modeling}
\label{fig:method}
\end{figure*}
According to the previous ideas, multiple modeling schemes will be generated under various combinations. The representative schemes are as follows.
\begin{itemize}
\item Minimalist base: directly model the completion order to obtain the elasticity curve.
(Not split into two-stage behavior sequence, nor separate modeling of natural quantity and incremental quantity)
\item Dual-task modeling: natural quantity and incremental quantity are built separately.
The two corresponding models can be placed in one network or built separately.
\item Sequence modeling: Completed orders equal to attendance rate multiplied by completed orders after attendance.
Elasticity curves are modeled for both attendance rate and completed orders after attendance.
\item Sequence and dual-task modeling: Completed orders equal to attendance rate multiplied by completed orders after attendance, and natural quantity and incremental quantity are built separately.
This model is the most complex and has the best effect. We call this as the Deep Monotonic Modeling of Multiple Causal Effects (MMCE).
\end{itemize}
\subsection{3.4 MMCE}
The network of MMCE is shown in Figure 5. The loss function is different for the blank group data and the non-blank group data.
\begin{figure*}[h]
\centering
\noindent \includegraphics[scale=0.54]{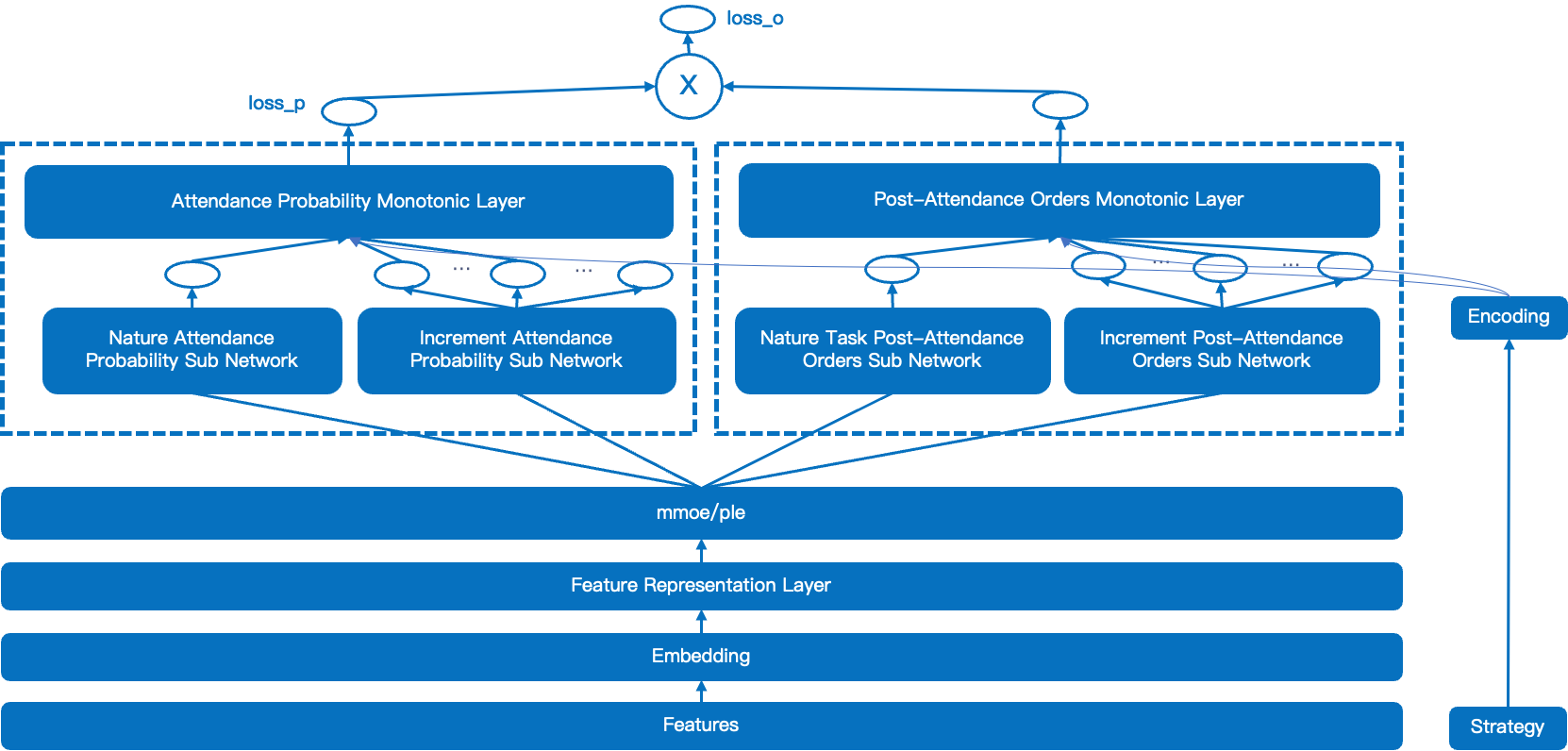}
\caption{MMCE}
\label{fig:method}
\end{figure*}
$$\mathop{{I}}\nolimits_{{v}}= a *loss\_p(t,x) + b *loss\_o(t,x)$$
First, the network is trained using the blank group data to obtain the results of the natural network. The incremental network does not update parameters and the results are not used.
Then the network is trained using non-blank group data to obtain the results of the incremental network, and the natural network does not update parameters.
\section{4 Experiments}
\subsection{4.1 Experimental Setup}
\textbf{Datasets} We use real request data sets for training and testing. The data sets are collected by a commercial company and anonymized according to the privacy policy. A total of more than 16 million request data were collected. About 14 million of them were used as training data sets, about 1 million were used as valid data sets, and about 1 million were used as test data sets. The data includes dozens of incentives, which you can understand as any value within a specific numerical range with one decimal place.

\textbf{Compared Models} We compared various methods. In order to demonstrate the superiority of this mechanism, we also implemented multiple methods.
\begin{itemize}
\item MMCE\_1:Build monotonicity separately.
\item MMCE\_2:Build two monotonies in one network.
\item MMCE\_3:Natural network is not separated from incremental network. Build two monotonies in one network.
\end{itemize}
Although There are many methods, such as those summarized in EFIN~\citep{liu2023explicit}, there are only a limited number of methods that directly support continuous treatment without any changes, such as DRNet~\citep{schwab2020learning} and VCNet~\citep{nie2021vcnet}. This article selects two of these results for comparison.

\textbf{Evaluation Metrics} We use cognitive testing to analyze the effects and propose under what conditions observational data can be considered as quantitative evaluation data set.
\begin{itemize}
\item \textbf{Qualitative Assessment} The method of cognitive testing is strongly related to the specific business, and the priors for different businesses will be different. This article explains the application process of this idea.
\begin{itemize}
\item Monotonicity. As the incentive increases, the response result becomes larger. Define the indicator to measure monotonicity. If the order of the result is the same as the order of the treatment, the numerator is added by 1. The denominator is the number of treatments.
\item Difference in stratification. High-ability riders  have higher natural results and smaller increments. Define the indicator to measure the difference in stratification, as shown in the Figure 6 below. If the curve of a low-ability rider is above all the high-ability riders, the numerator is added by 1. The denominator is the number of rider stratifications.
\begin{figure*}[h]
\centering
\noindent \includegraphics[scale=0.34]{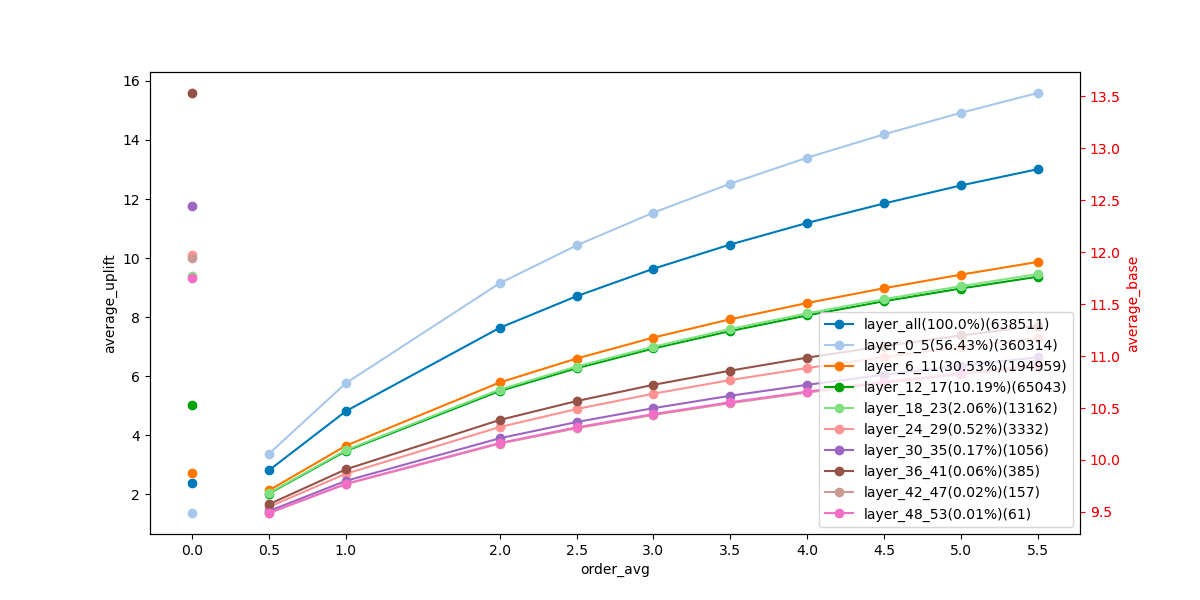}
\caption{Price response curves for different stratification of riders}
\label{fig:method}
\end{figure*}
\item Diminishing marginal Effect. As costs rise, ROI becomes lower and lower. As shown in the Figure 7 below, define the indicator to measure diminishing marginal effect. if the  ROI of the lowest treatment is the largest, the numerator is added by 1. The denominator is the number of rider stratifications.
\begin{figure*}[h]
\centering
\noindent \includegraphics[scale=0.34]{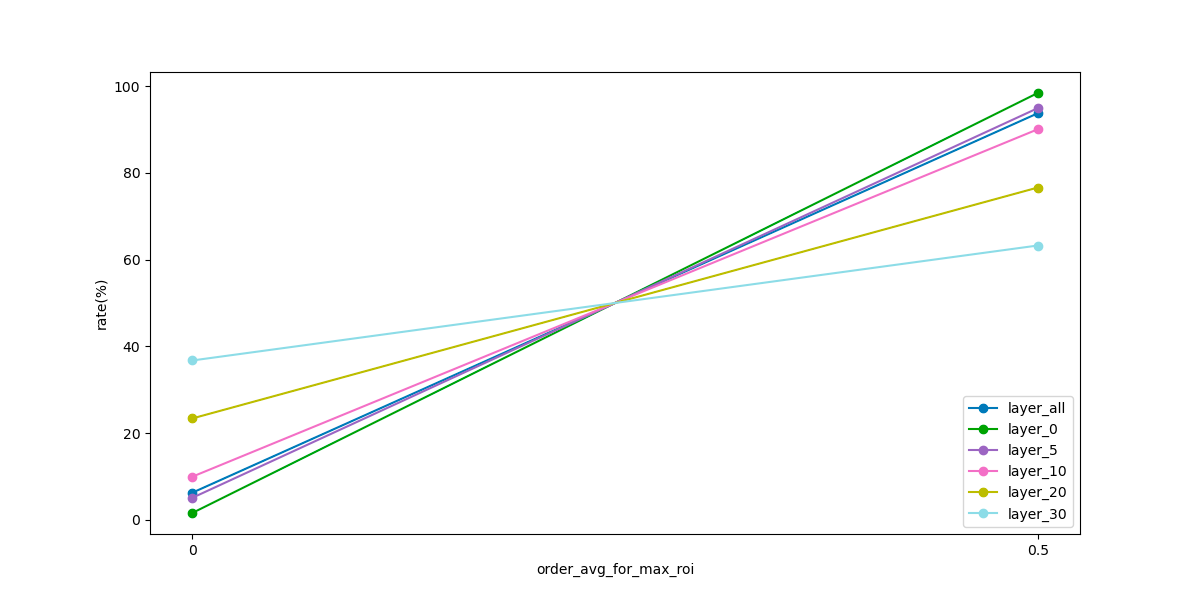}
\caption{Maximum ROI for different stratification of riders}
\label{fig:method}
\end{figure*}
\end{itemize}

\item \textbf{Quantitative indicators} Without RCT data, it is impossible to conduct an evaluation. Can observational data be used to assist in the evaluation? RCT data can be used for evaluation because it naturally meets the three major assumptions in the potential causal framework. Through business cognition, we can confirm that the data meets assumption three, and check whether assumption two is met for important features. Assumption one is generally difficult to test. We found that when the observational data better meets the cognitive testing in qualitative evaluation at a macro level, the evaluation results obtained using the observational data have certain reference value. Because it is basically positively correlated with the results in qualitative evaluation. The conventional gini score can also be tried for evaluation.
\begin{itemize}
\item assumption one is Conditional Independence Assumption(CIA). when given variable $X$, the relationship between intervention variable $T$ and outcome variable $Y$ is independent. The formula is: $$(Y_i(0),Y_i(1)) \perp T_i|X_i$$ 
\item assumption two is Positivity Assumption(Positivity). T=0 and T=1 have results for any X. The formula is: $$ 0<P_r(T_i=1|X_i)<1, \forall\ X_i \in X $$
\item assumption three is Stable Unit Treatment Value Assumption(SUTVA). The results of any individual will not change due to other individuals. For an individual, the same intervention will not lead to different results.
\end{itemize}
\end{itemize}
\subsection{4.2 Offline  Results}
The effect analysis was performed using cognitive testing. Table 1 shows the results of the quantitative analysis of different priors. It is obvious that the effect of the MMCE framework is obvious, and MMCE-3 also shows better utilization of the blank group data.
\begin{table*}[h]
\centering
\caption{Quantitative results for different models with different priors}
\label{tab:evaluation}
\begin{tabular}{|c|c|c|c|}
\hline
\textbf{model}  & \textbf{monotonicity} & \textbf{difference in stratification} & \textbf{marginal Effect} \\ \hline
\textbf{DRNet}  & 0.4                   & 0.56                         & 0.7                       \\ \hline
\textbf{VCNet}  & 0.6                   & 0.56                         & 0.8                       \\ \hline
\textbf{MMCE-1} & 1                     & 0.67                         & 1                         \\ \hline
\textbf{MMCE-2} & 1                     & 0.67                         & 1                         \\ \hline
\textbf{MMCE-3} & 1                     & 0.78                         & 1                         \\ \hline
\end{tabular}
\end{table*}
The importance of finding reasonable evaluation data for modeling is self-evident. There are many methods in the industry to construct data, such as match. According to the previous analysis, the gini score was evaluated using observation data that met the requirements, and the MMCE architecture also shows better result.
\begin{table*}[h]
\centering
\caption{Evaluated the gini score}
\label{tab:evaluation}
\begin{tabular}{|c|c|}
\hline
\textbf{model}  & \textbf{gini score} \\ \hline
\textbf{DRNet}  & 0.2                 \\
\textbf{VCNet}  & 0.3                 \\
\textbf{MMCE-1} & 0.46                \\ \hline
\textbf{MMCE-2} & 0.47                \\
\textbf{MMCE-3} & 0.48                \\ \hline
\end{tabular}
\end{table*}
\subsection{4.3 Offline  Results}
Under the guidance of the MMCE framework concept, we implemented a simpler version in the early stage. In order to verify the effectiveness of the method, we further conducted A/B testing in a real online environment. In the A/B test, we first randomly divided all candidates into two groups (experiment group and control group). In the case of budget balancing, the experiment group allocated incentives to users according to the estimated response model. After the experiment, the ROI of the experiment group increased by 12\% compared with the control group.
\section{5 Conclusion and Discussion}
Our main contributions:
\begin{itemize}
\item The first to propose a framework for multi-stage causal modeling based on large-scale observational data to generate multiple causal effects at the same time.
\item The network structure is highly scalable.
\item This paper summarizes three basic priors to illustrate the necessity and feasibility of qualitative evaluation of cognitive testing. At the same time, this paper innovatively proposes under what conditions observational data can be considered as an evaluation data set.
\end{itemize}
Modeling and evaluation of observational data are very challenging in themselves, and every small step forward is very valuable. Obviously, the network structure and evaluation in this article rely on business cognition, which is still satisfied in some cases in the growth field. But some cognition is not satisfied in some scenarios, which is the application limitation of this article.

\section{6 Acknowledgments}
I would like to thank all the reviewers and those who have given me help and encouragement in this work.

\section{appendix}
\begin{itemize}
\item \textbf{Linear Function}
$$y= a * log\ (x+1) +b$$
A and b are the parameters to be learned. Assuming a=0.7 and b=1, the relationship between x and y is shown in the Figure 8.
\begin{figure}
    \centering
    \includegraphics[width=0.5\linewidth]{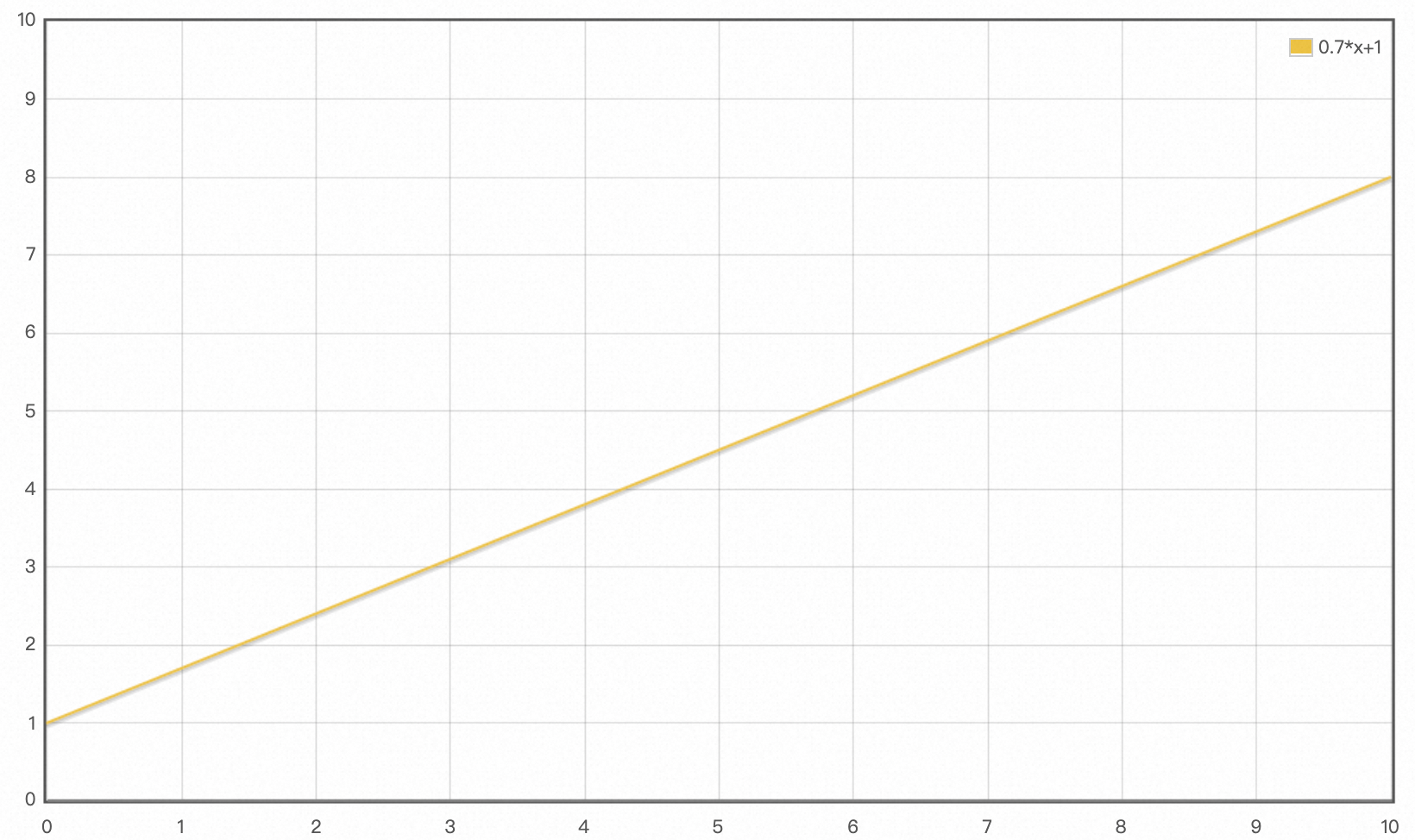}
    \caption{Linear function}
    \label{fig:enter-label}
\end{figure}

\item \textbf{Logarithmic function}
$$y= a * log\ (x+1) +b$$
A and b are the parameters to be learned. Assuming a=2 and b=1, the relationship between x and y is shown in the Figure 9.
\begin{figure}
    \centering
    \includegraphics[width=0.5\linewidth]{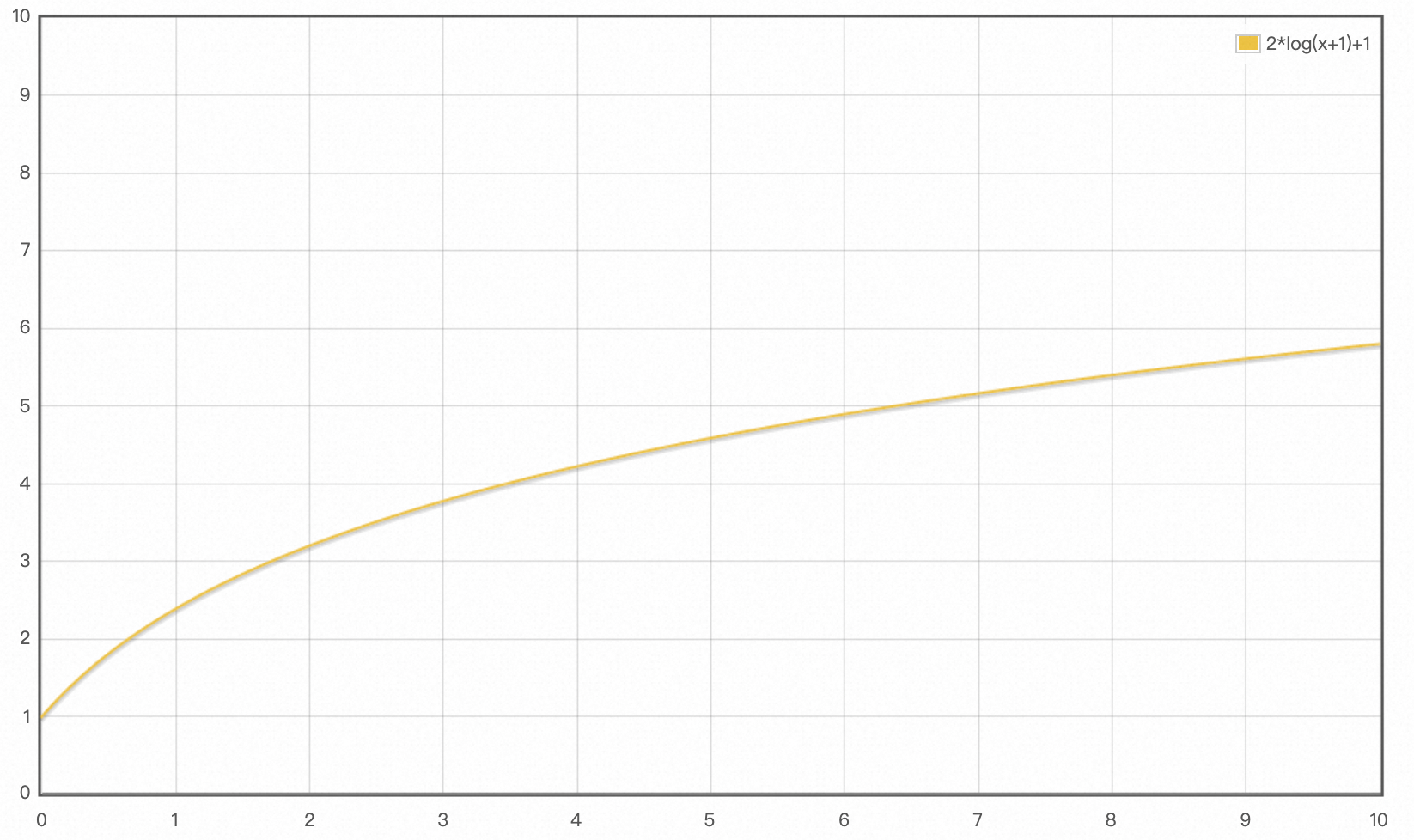}
    \caption{Logarithmic function}
    \label{fig:enter-label}
\end{figure}
\item \textbf{S-shaped function}
$$y = \frac{D}{1+e^{-a*x + b }}$$
$D$ is the maximum value of $y$. When D=5 and a and b are the parameters to be learned. Assuming a=1 and b=1, the relationship between x and y is shown in the Figure 10.
\begin{figure}
    \centering
    \includegraphics[width=0.5\linewidth]{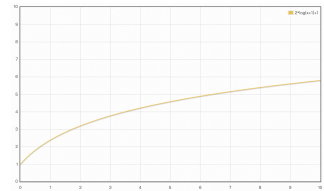}
    \caption{S-shaped function}
    \label{fig:enter-label}
\end{figure}

\item \textbf{isotonic encoding lr}
$$y =  w^T * encoding(x) = \sum_{i=0}^{N} x_i * w_i$$
$$encoding(x) = [x_0,x_1,x_2,...,x_N]$$
$$\mathop{{x}}\nolimits_{{i}}= \left\{ \begin{array}{*{20}{l}}
{1,}&{if\ i<= x}\\
{0,}&{if\ i> x}
\end{array}\right. $$
$$w=[w_0,w_1,w_2,...,w_N],w_i \ge 0$$
N is the maximum value of x. When N=9, w is the parameter vector to be learned, which is first fixed as a known vector. The calculation of each value is shown in Table 3.
\begin{table*}[h]
\centering
\caption{Calculation process demonstration}
\label{tab:evaluation}
\begin{tabular}{|c|c|c|c|}
\hline
\textbf{x} & \textbf{encoding(x)}               & \textbf{w}                              & \textbf{y}   \\ \hline
0           & {[}1,0,0,0,0,0,0,0,0,0{]} & {[}2,1,1.5,1,0.7,0.7,0.7,0.5,0.5,0.2{]} & 2   \\ \hline
1           & {[}1,1,0,0,0,0,0,0,0,0{]} & {[}2,1,1.5,1,0.7,0.7,0.7,0.5,0.5,0.2{]} & 3   \\ \hline
2           & {[}1,1,1,0,0,0,0,0,0,0{]} & {[}2,1,1.5,1,0.7,0.7,0.7,0.5,0.5,0.2{]} & 4.5 \\ \hline
3           & {[}1,1,1,1,0,0,0,0,0,0{]} & {[}2,1,1.5,1,0.7,0.7,0.7,0.5,0.5,0.2{]} & 5.5 \\ \hline
4           & {[}1,1,1,1,1,0,0,0,0,0{]} & {[}2,1,1.5,1,0.7,0.7,0.7,0.5,0.5,0.2{]} & 6.2 \\ \hline
5           & {[}1,1,1,1,1,1,0,0,0,0{]} & {[}2,1,1.5,1,0.7,0.7,0.7,0.5,0.5,0.2{]} & 6.9 \\ \hline
6           & {[}1,1,1,1,1,1,1,0,0,0{]} & {[}2,1,1.5,1,0.7,0.7,0.7,0.5,0.5,0.2{]} & 7.6 \\ \hline
7           & {[}1,1,1,1,1,1,1,1,0,0{]} & {[}2,1,1.5,1,0.7,0.7,0.7,0.5,0.5,0.2{]} & 8.1 \\ \hline
8           & {[}1,1,1,1,1,1,1,1,1,0{]} & {[}2,1,1.5,1,0.7,0.7,0.7,0.5,0.5,0.2{]} & 8.6 \\ \hline
9           & {[}1,1,1,1,1,1,1,1,1,1{]} & {[}2,1,1.5,1,0.7,0.7,0.7,0.5,0.5,0.2{]} & 8.8 \\ \hline
\end{tabular}
\end{table*}

\end{itemize}

\newpage
\bibliographystyle{aaai} 
\bibliography{ref_new}

\end{document}